\documentclass[letterpaper]{article} 
\usepackage{aaai25}  
\usepackage{times}  
\usepackage{helvet}  
\usepackage{courier}  
\usepackage[hyphens]{url}  
\usepackage{graphicx} 
\usepackage{natbib}  
\usepackage{caption} 
\usepackage{algorithm}
\usepackage{algorithmic}
\usepackage{amsmath}
\usepackage{booktabs}
\usepackage{amssymb}
\usepackage{graphicx} 

\usepackage{subfigure}
\usepackage[T1]{fontenc}
\usepackage{newfloat}
\usepackage{listings}
\DeclareCaptionStyle{ruled}{labelfont=normalfont,labelsep=colon,strut=off} 
\floatstyle{ruled}
\newfloat{listing}{tb}{lst}{}
\floatname{listing}{Listing}
\title{FAP-CD: Fairness-Driven Age-Friendly Community Planning via Conditional Diffusion Generation}
\author{
    Jinlin Li\textsuperscript{\rm 1}, Xintong Li\textsuperscript{\rm 2}, Xiao Zhou\textsuperscript{\rm 1 }\thanks{Corresponding author}
}
\affiliations{
     \textsuperscript{\rm 1} Gaoling School of Artificial Intelligence, Renmin University of China, China\\
    \textsuperscript{\rm 2} School of Statistics, Renmin University of China, China\\
    jinlin2021@ruc.edu.cn, lixintong@ruc.edu.cn, xiaozhou@ruc.edu.cn

}

\usepackage{bibentry}

\begin{document}

\maketitle

\begin{abstract}

As global populations age rapidly, incorporating age-specific considerations into urban planning has become essential to addressing the urgent demand for age-friendly built environments and ensuring sustainable urban development. However, current practices often overlook these considerations, resulting in inadequate and unevenly distributed elderly services in cities. There is a pressing need for equitable and optimized urban renewal strategies to support effective age-friendly planning. To address this challenge, we propose a novel framework, \textbf{F}airness-driven \textbf{A}ge-friendly community \textbf{P}lanning via \textbf{C}onditional \textbf{D}iffusion generation (FAP-CD). FAP-CD leverages a conditioned graph denoising diffusion probabilistic model to learn the joint probability distribution of aging facilities and their spatial relationships at a fine-grained regional level. Our framework generates optimized facility distributions by iteratively refining noisy graphs, conditioned on the needs of the elderly during the diffusion process. Key innovations include a demand-fairness pre-training module that integrates community demand features and facility characteristics using an attention mechanism and min-max optimization, ensuring equitable service distribution across regions. Additionally, a discrete graph structure captures walkable accessibility within regional road networks, guiding model sampling. To enhance information integration, we design a graph denoising network with an attribute augmentation module and a hybrid graph message aggregation module, combining local and global node and edge information. Empirical results across multiple metrics demonstrate the effectiveness of FAP-CD in balancing age-friendly needs with regional equity, achieving an average improvement of 41\% over competitive baseline models.

\end{abstract}

\begin{links}
\link{Code}{https://github.com/jinlin2021/FAP_CD}
\end{links}

\section{Introduction}

Over the past two decades, global population aging has accelerated. By 2050, individuals aged 65 and older are projected to comprise 38\% of the world’s population~\cite{rudnicka2020world}, posing new demands on urban infrastructure and services. To address these challenges, the World Health Organization (WHO) introduced the concept of Age-friendly Cities in 2007~\cite{world2007global}, highlighting the need for developing Age-friendly Communities (AFCs) to support aging populations~\cite{menec2011conceptualizing}.
AFCs aim to improve the quality of life and social participation of elderly residents through enhanced community services. Research further shows that most elderly individuals prefer to age in residential communities rather than rely on institutional care services~\cite{sun2023application}. In this context, studying and optimizing the distribution of aging-related facilities and services within communities is essential for the effective development of age-friendly cities.

\begin{figure}[!tb]
\includegraphics[width=\linewidth]{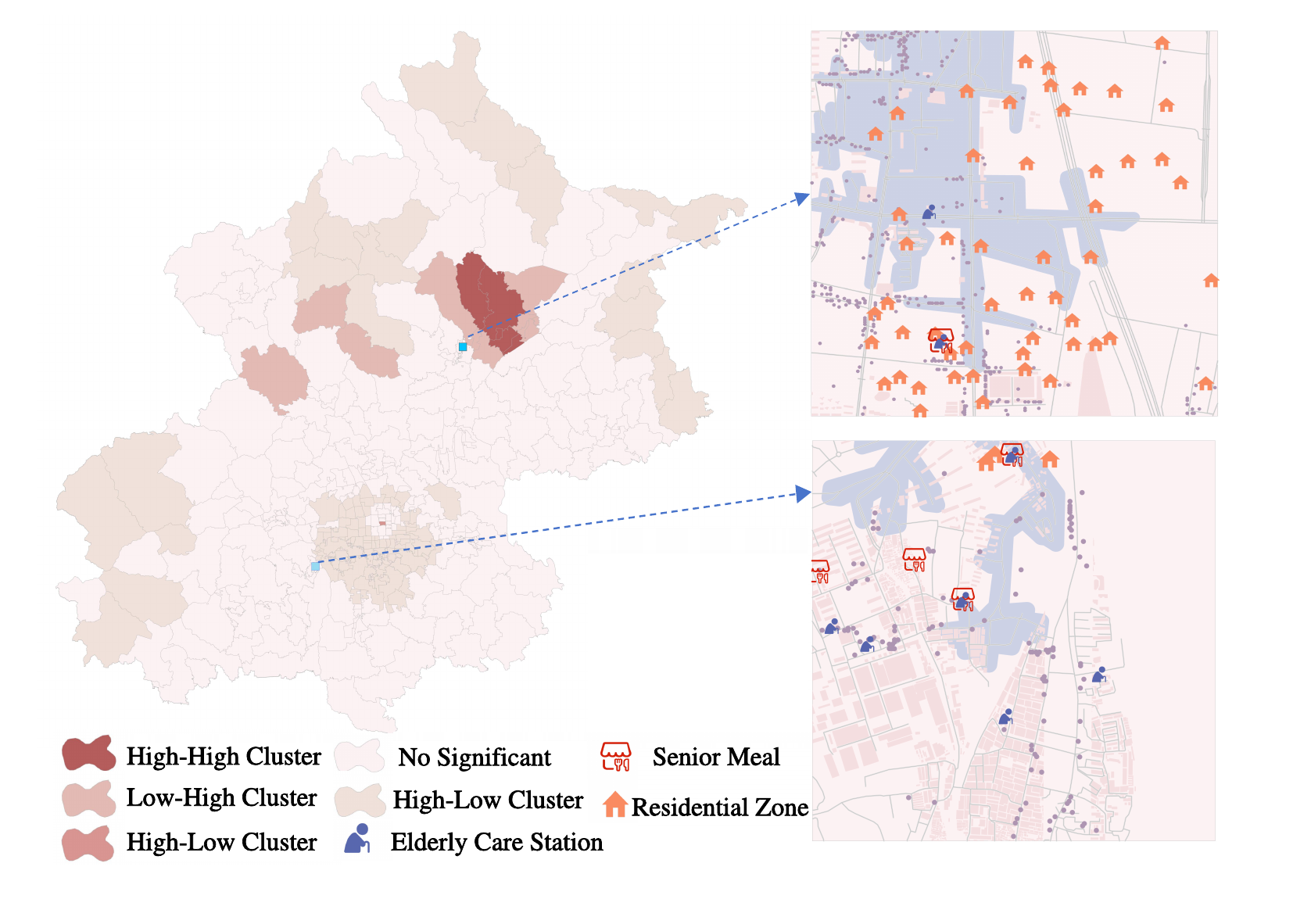}
\caption{Local Moran's Index analysis of the distribution of age-related facilities in Beijing, illustrating spatial disparities in resource allocation. Two regions with contrasting resource levels are magnified, with blue-shaded zones indicating 15-minute walkable areas around facilities.}
\label{fig:Moran}
\end{figure}

Unfortunately, even in some of the most advanced and developed modern metropolises, the level of age-friendly infrastructure remains far from adequate~\cite{wang2017applying, xie2018age}. For instance, in Beijing, although 22.6\% of the population was aged 60 and above by 2023, significant deficiencies in elderly care facilities persist. Our analysis, leveraging the 15-minute Walkability metric~\cite{moreno2021introducing}—which evaluates resource accessibility and captures the ease of access to community-based elderly services—shows that only 40.85\% of communities have senior care centers within a 15-minute walking distance, and merely 24.51\% offer meal assistance services for the elderly. Furthermore, an analysis using Local Moran's Index reveals significant spatial inequalities in the distribution of elderly care facilities across Beijing’s neighborhoods, as illustrated in Figure~\ref{fig:Moran}. More critically, as the aging population grows, demands for medical, recreational, and educational facilities become increasingly diverse and urgent~\cite{yung2016social}, yet current facility allocations fail to meet these needs or support AFCs development and regeneration.

The planning of age-friendly communities is crucial for achieving active aging among community seniors and establishing sustainable communities. The core task is to create an optimized spatial layout that reflects the needs of the elderly in urban areas~\cite{sun2023application}. Traditional large-scale urban planning, from initial sketches to detailed designs, is not only complex and time-consuming but also incurs substantial costs. In today's rapidly digitizing world, fueled by urban big data and advancements in artificial intelligence (AI), automated spatial layout design has not only become feasible but is also emerging as a prevailing trend.

In recent years, a few studies have attempted to generate planning schemes automatically through deep generative techniques. For instance,~\citeauthor{wang2020reimagining}\shortcite{wang2020reimagining} proposed a generative model for land use allocation, utilizing geographic and human mobility data to build spatial graph representations and training a generative adversarial network with positive and negative sample contrasts.~\citeauthor{wang2023human}\shortcite{wang2023human} developed an instruction-based hierarchical planner (IHPlanner) that captures the spatial hierarchy and planning dependencies between urban functional zones and land use allocation using human instructions and the surrounding environment, generating land use configurations for vacant areas.~\citeauthor{zheng2023spatial}\shortcite{zheng2023spatial} proposed a reinforcement learning (RL)-based urban planning model, constructing an urban adjacency graph with urban geographic elements (blocks, roads) as nodes and framing urban planning as a sequential decision-making problem on the graph. Additionally, some studies~\cite{ye2022masterplangan, eskandar2023urban} have viewed urban areas as two-dimensional images, treating scheme generation as an image generation task.

Although these methods have advanced the field of automated urban planning scheme generation, they are limited in scope: either simulating urban development patterns at a macro level while neglecting detailed land use and building layouts, or coarsely updating land allocation without incorporating the urban multimodal data~\cite{yong2024musecl,xu2024cgap} of target communities.  
Moreover, no algorithms currently exist for generating planning schemes specifically tailored to age-friendly urban renewal. Addressing this critical technical gap is an urgent priority.

Recently, diffusion models~\cite{ho2020denoising, song2020denoising} have made significant advancements in the field of computer vision, becoming the most advanced deep generative models, surpassing other generative models such as Variational Autoencoders (VAEs)~\cite{kingma2013auto} and Generative Adversarial Networks (GANs)~\cite{goodfellow2020generative}. Diffusion models approximate complex data distributions by gradually removing small amounts of noise from a simple distribution. The refined denoising process enables these models to effectively fit complex distributions and generate high-quality data. In this study, we propose a fairness-driven age-friendly community planning method (FAP-CD) based on the denoising diffusion model guided by stochastic differential equations (SDEs). Leveraging conditional diffusion generation, this approach frames the planning problem as a facility graph generation task. It utilizes complex walking graphs to capture spatial relationships among regional facilities, aiming to better meet the needs of elderly residents while promoting fairness across regions.

More specifically, we deploy a unified graph noise prediction model that intricately interacts with the node and edge representations in both real-valued matrices and discrete graph structures. Initially, we develop a fair-demand pre-training module, utilizing min-max optimization~\cite{pardalos2008pareto} to derive spatial region representations that achieve maximum minimum entropy~\cite{han2021max,razaviyayn2013linear}. These representations are meticulously crafted to reflect residential demands within the region and to promote equitable characteristics across communities. Subsequently, we guide the diffusion process by carefully adjusting the mean of the introduced Gaussian noise. During the denoising phase, we implement a hybrid graph noise prediction model, which comprises a message-passing layer on the discrete graph and a graph transformer layer. The message-passing layer is responsible for modeling the dependencies among the edges of neighboring nodes, while the transformer layer is dedicated to extracting and disseminating global information. Furthermore, we utilize a straightforward walking capability graph to support conditional discrete graph denoising and employ m-step random walk matrices, derived from the discrete adjacency matrix, to enhance the representation learning of nodes and edges. 

Our main contributions are summarized as follows:
\begin{itemize}
\item  We propose an efficient conditional discrete graph diffusion framework. To the best of our knowledge, this is the first study on generating AFCs planning that consider both elderly needs and equity.

\item We design a novel graph denoising diffusion network conditioned on fair-demand embedding and simple discrete structure to ensure practicality and fairness of the spatial layout in the generated grid layouts.

\item Extensive experiments on real-world multi-modal urban data demonstrate that FAP-CD delivers superior performance and significantly outperforms contemporary state-of-the-art methods. Additionally, we provide in-depth analyses to highlight the superiority of our framework.

\end{itemize}

\section{Preliminaries}

\subsection{Problem Statement}

Given a city area divided into an H × W grid map based on latitude and longitude, with each grid of size $d \times d\,m^2$, the grid regions are denoted as $R = \{r_1, r_2, \ldots, r_k\}$, where $k$ is the number of grid regions in the city.
In our study, \(d\) is set to 2 km, and each grid encompasses the following aspects:

\noindent\textbf{Urban attributes.}
Urban attributes are the inherent social and geographic features of urban regions. A specific type of regional attribute can be denoted as $A \in \mathbb{R}^{N \times f_a}$, where $f_a$ is the region attribute dimensions. In our work, multiple region attributes, like demographics and socio-economic indicators, are considered.

\noindent\textbf{Walking graph.}
For each grid region $r_i$, we construct an undirected weighted graph, denoted as $\boldsymbol{G} = \{\boldsymbol{N}, \boldsymbol{E}, \boldsymbol{A}\}$, where $\boldsymbol{N}$, $\boldsymbol{E}$ are the node set and edge set, respectively. $\boldsymbol{A}\in\{0,1\}^{N\times N}$ is the adjacency matrix to describe the walkable accessibility between every two nodes in the region. $\boldsymbol{N} \in \mathbb{R}^{D \times f_d}$, where $D$ denotes the total number of facilities and $f_d$ dictates the distinct categories of facilities within the grids $R$. Each category is encoded using a one-hot representation. Additionally, textual modal descriptions of each facility are captured as grid features $\boldsymbol{F} \in \mathbb{R}^{N \times d}$.

\noindent\textbf{Grids region demand.}
This includes the current status of facility distribution on each grid region, based on Beijing’s standards for facilities per thousand residents\footnote{http://www.cacp.org.cn} and the regulations for age-appropriate amenities\footnote{https://www.beijing.gov.cn/zhengce/zhengcefagui/}, we evaluate the existing facility on each grid. We classify the state of area facilities into four categories: no supply, under supplied, appropriately supplied, and oversupplied. We standardize each demand representation $\boldsymbol{D}\in\{0,1,2,3\}^{N\times N}$(i.e., high configuration, proper fit, low configuration, under allocation).

\noindent\textbf{AFCs Planning Generation.}
Given a set of grid regions $R = \{r_1, r_2, \ldots, r_k\}$, each characterized by grid demand $\boldsymbol{D}$, urban region attribute $A$ and grid feature $\boldsymbol{F}$, our goal is to generate the AFCs planning that addresses equitable demands across various urban areas in different time intervals.

\subsection{Preliminaries on Diffusion Process}

Diffusion models are a class of generative models that generate samples by learning a distribution that approximates a data distribution. The first step in constructing diffusion models~\cite{ho2020denoising, song2020denoising} is to define a forward diffusion process. Assuming a continuous random variable $\boldsymbol{x_0}\in\mathbb{R}^d$, the diffusion models consist of a forward process that gradually add noise to perturbs $\boldsymbol{x_0}$ until the output distribution becomes a known prior distribution within the specified time range \(t\) in the interval \([0, T]\), we have a Gaussian transition kernel as follows:
\begin{eqnarray}
\label{eq:equation0}
q_{0t}(\boldsymbol{x}_t|\boldsymbol{x}_0)=\mathcal{N}(\boldsymbol{x}_t|\alpha_t\boldsymbol{x}_0,\sigma_t^2\boldsymbol{I}) ,
\end{eqnarray}
where $\alpha_t$ and $\sigma_t$ are time-dependent differentiable functions, usually chosen to ensure that the signal-to-noise ratio $\frac{\alpha_t^2}{\sigma_t^2}$ decreases. This ensures that $q_T(x_T) \approx \mathcal{N}(0, I)$, resulting in a tractable prior distribution with a low signal-to-noise ratio that is convenient for sampling. The forward process can be represented by the following SDE:
\begin{equation}
    \begin{aligned}
    \label{eq:equation1}
    & \mathrm{d}\boldsymbol{x}_t={f(\boldsymbol{x}_t, t)}\mathrm{d}t+{g(t)}\mathrm{d}\boldsymbol{w}_t ,
    \end{aligned}
\end{equation}
where $f(\cdot,t)$ is the drift coefficient, $g(t)$ is the scalar diffusion terms (a.k.a.\ noise schedule function), and $\boldsymbol{w}_t$ is a standard Brownian motions on $f(\cdot,t)$. The reverse-time SDEs from time T to 0 is denoted as:
\begin{footnotesize}
\begin{equation}
\begin{aligned}
\label{eq:equation2}
\mathrm{d}\bar{\boldsymbol{x}}_t = [f(t)\boldsymbol{x}_t-g^2(t)\nabla\log q_t(\boldsymbol{x}_t)]\mathrm{d}t+ g(t)\mathrm{d}\bar{\boldsymbol{w}}_t ,
\end{aligned}
\end{equation}
\end{footnotesize}where $\nabla\log q_t(\boldsymbol{x}_t)$ is the score function and $\mathrm{d}\bar{\boldsymbol{w}}_t$ is the reverse-time standard Wiener process. By learning to reverse this process, the diffusion model generates new samples from the data distribution. 

In this work, to address the scarcity of elderly-friendly facilities and spatial inequity within contemporary urban structures, we adopt a diffusion model based on graph SDEs. Specifically, we abstract the spatial and facility relationships within grid region into a graphical representation, constructing a graph $\boldsymbol{G}=(\boldsymbol{X},\boldsymbol{A})$ composed of N nodes with features $X{\in}\mathbb{R}^{N\times F}$ encoded using one-hot encoding to denote the types of facilities, and the adjacency matrix $\mathbf{A} \in \mathbb{R}^{N \times N}$ (i.e., $A_{i,j} = 1$ if the walking time between node $i$ and $j$ is within 15 minutes).

\begin{figure}[!tb]
\includegraphics[width=\linewidth]{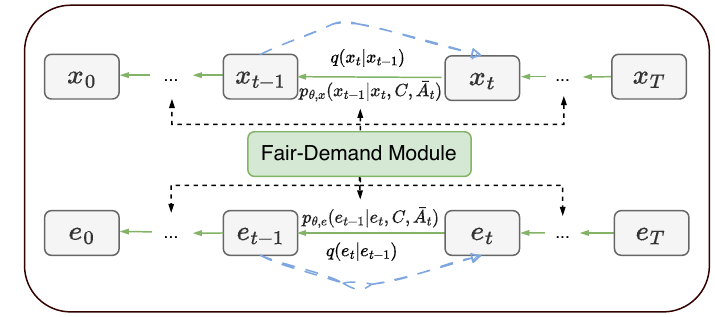}
\caption{Framework overview of the proposed FAP-CD.} 
\label{fig:graph denoising network}
\end{figure}

\section{Methodology}

\begin{figure*}[!t]
	\centering
        \includegraphics[width=0.95\textwidth]{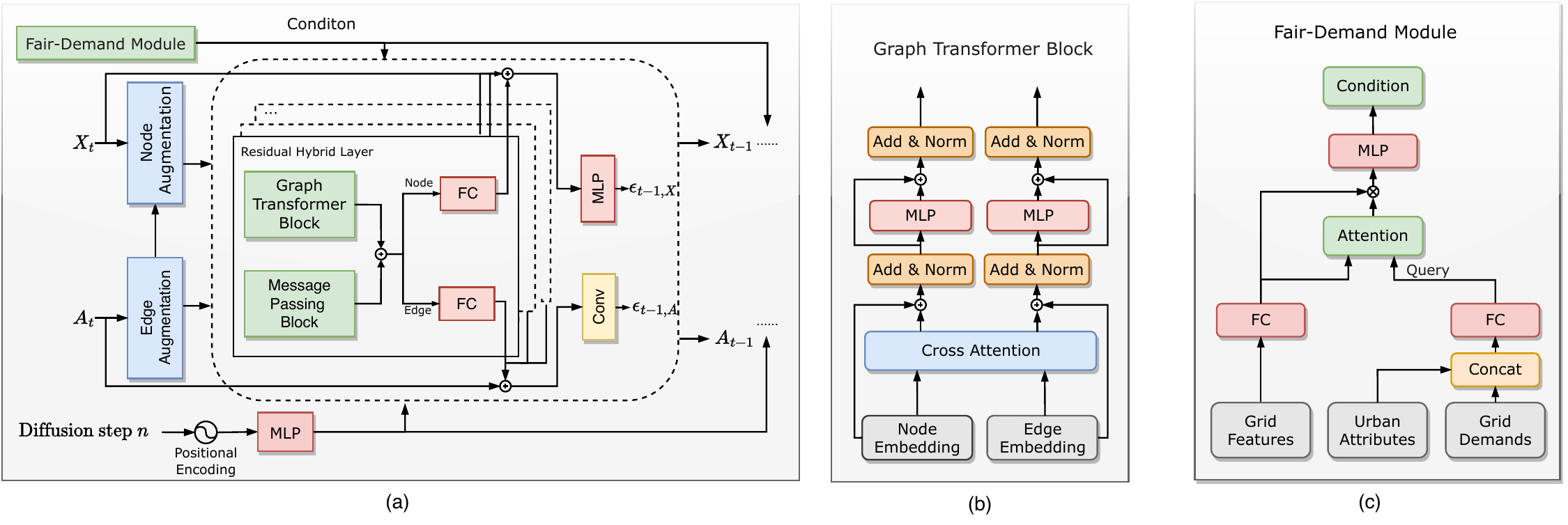}
	\caption{The architecture of (a) graph denoising network, (b) graph transfomer block, and (c) fair-demand module.}
	\label{fig:framework}
	\vspace{-2ex}
\end{figure*}
Figure~\ref{fig:graph denoising network} illustrates the overall framework of FAP-CD, which can be divided into two stages: forward process and reverse process with conditional graph denoising. The fair-demand module learns to ensure a valid fair representation optimized by the min-max principle across regions, which serves as the condition supporting the graph denoising diffusion process.

\subsection{Fair-demand Module} 

To generate a target optimized spatial distribution map of facilities from noise in vector space, we design an attention mechanism-based fair-demand module as shown in Figure~\ref{fig:framework} (c). The denoising process utilizes the region-specific max-min fairness demand embeddings, which are learned through the fair-demand module, to guide the AFCs generation. Specifically, for each region \( r_i \), let $\bar{D_i}$ denotes the representation obtained by concatenating the urban attributes and demand representations, then use a linear projection to obtain the query vector \( Q = \bar{D}_i W_q \). Subsequently, grid features are mapped through a fully connected layer to the same vector space to get the key vector \( K = F_i W_k \) and the value vector \( V = F_i W_v \). The importance of region demand representations and the output are calculated as follows:
\begin{align}
    & \alpha_{i}=softmax(\frac{QF_{i}^{\top}}{\sqrt{d_{h}}}),
    \\
    & H_{i}=\alpha_{s}\mathrm{V}W_{H}, 
\end{align}where $W_{O}$ is the output projection matrix.

In this study, we advance beyond simply concatenating various conditional embeddings for control conditions. We focus on detailed modeling of demand interactions among facilities and the semantic characteristics of residential areas, while also addressing inter-regional fairness. Our refined conditional embedding, represented as \(C_{i} = H_{i} \circ F_i\), empowers the model to adaptively leverage these representations for more nuanced and effective outcomes.

To generate a distribution of service facilities that meets the requirements of age-friendliness and spatial fairness, we develop a pre-trained model optimized based on the principle of max-min entropy. This framework has been widely applied across various optimization and control domains~\cite{liu2013max, zhang2018mitigating, wai2019variance}. Our model is trained by minimizing the loss function:
\begin{equation}
\boldsymbol{L} = \frac{1}{min(-\bar{p}_i* log(\bar{p}_i)) +1 } ,i \in M,
\end{equation}
where $ \bar{p}_i = \sum_{i=1}^{n} p_{ij} $, represents the normalized probability distribution, while $ p_{ij} = 0 $ if $ j \in R_i $. The term $ p_{ij} $ is the softmax output of $ O_i $, $n$ is the batch size, $ R_i $ denotes the index of the residence within the grid region. $ M $ represents the maximum number of facilities per region.

\subsection{Discrete Graph Diffusion Model}
In this work, we formulate the AFCs planning generation task in a conditional SDEs-based graph denoising framework. Inspired by~\cite{jo2022score,vignac2022digress} which treats (\(X, A\)) as a whole and omit the intrinsic graph structure. we diffuse separately on each node and edge feature of the constructed graph $\boldsymbol{G}=(\boldsymbol{X},\boldsymbol{A})$. We apply the variance-preserving SDE introduced by~\cite{song2020score}, which in its discrete-time manifestation adopts the framework of a denoising diffusion probabilistic model~\cite{ho2020denoising}. Similarly to diffusion models for images, which apply noise independently on each pixel, we diffuse separately on each node and edge feature. Roughly speaking, the forward diffusion process for each graph sample can be described by the SDE with $t\in[0, T]$ as:
\begin{equation}
    \begin{aligned}
    & \mathrm{d}\boldsymbol{G}_t={f(\boldsymbol{G}_t, t)}\mathrm{d}t+{g(t)}\mathrm{d}(\boldsymbol{w}_{X_t} ,\boldsymbol{w}_{A_t}),
    \end{aligned}
\end{equation}
where $\boldsymbol{w}_{X_t}$ and $\boldsymbol{w}_{A_t}$ are independent standard Wiener processes for the node and edge, respectively. This forward SDE shares the same transition distribution as in Eq.~\ref{eq:equation0} to conveniently sample Gaussian noise $\boldsymbol{G}_t = \alpha_t \boldsymbol{G}_0 + \sigma_t \boldsymbol{\epsilon}_G$ at any time $t$. For undirected graphs, noise is applied only to the upper triangular part of $\boldsymbol{E}$ and then symmetrized to maintain the undirected nature of the graph.

In the reverse process with conditional denoising, FAP-CD defines the conditional reverse diffusion process $p_{\theta}(\boldsymbol{G}_{t-1} | \boldsymbol{G}_t, \boldsymbol{C}, \boldsymbol{\bar A})$, which performs iterative denoising from pure Gaussian noise to generate optimized spatial facilities in vector space. This process is conditioned on the region fair-demand embeddings $\boldsymbol{C}$ and a discrete adjacency matrix $\boldsymbol{\bar A}$, representing facility nodes within a 15-minute walk from residential areas. The corresponding reverse-time SDEs are given by:
\begin{equation}
\resizebox{0.9\columnwidth}{!}{$
\begin{aligned}
\mathrm{d}\boldsymbol{X}_t &= \left[f(t)\boldsymbol{X}_t - g^2(t)\nabla_{\boldsymbol{X}}\log q_t(\boldsymbol{X}_t, \boldsymbol{A}_t)\right]\mathrm{d}t + g(t)\mathrm{d}\bar{\boldsymbol{w}}_{X_t}, \\
\mathrm{d}\boldsymbol{A}_t &= \left[f(t)\boldsymbol{A}_t - g^2(t)\nabla_{\boldsymbol{A}}\log q_t(\boldsymbol{X}_t, \boldsymbol{A}_t)\right]\mathrm{d}t + g(t)\mathrm{d}\bar{\boldsymbol{w}}_{A_t}.
\end{aligned}
$}
\end{equation}

Then, we train two time-dependent score networks $\boldsymbol{\epsilon}_{\boldsymbol{\theta},\boldsymbol{X}}(\cdot)$ and $\boldsymbol{\epsilon}_{\boldsymbol{\theta},\boldsymbol{A}}(\cdot)$ i.e., the parametric score function, for estimating the score $\nabla_{\boldsymbol{X}}\log q_t(\boldsymbol{X}_t,\boldsymbol{A}_t)$ and $\nabla_{\boldsymbol{A}}\log q_t(\boldsymbol{X}_t,\boldsymbol{A}_t)$ for both $(X_t , A_t)$, separately. The denoising score matching objective~\cite{vincent2011connection} is modified for score estimation training as:
\begin{footnotesize}
\begin{equation}
	\begin{split}
    \min_{\boldsymbol{\theta}}\mathbb{E}_t\{\lambda(t)\mathbb{E}_{\boldsymbol{G}_0}\mathbb{E}_{\boldsymbol{G}_t|\boldsymbol{G}_0}[&||\boldsymbol{\epsilon}_{\boldsymbol{X}} -\boldsymbol{\epsilon}_{\boldsymbol{\theta},\boldsymbol{X}}(\boldsymbol{G}_t,\boldsymbol{C}, \boldsymbol{\bar A}_t,t)||_2^2+ \\& ||\boldsymbol{\epsilon}_{\boldsymbol{A}} -\boldsymbol{\epsilon}_{\boldsymbol{\theta},\boldsymbol{A}}(\boldsymbol{G}_t,\boldsymbol{C}, \boldsymbol{\bar A}_t,t)||_2^2]\}\mathrm{~},
	\end{split}
\end{equation}
\end{footnotesize}where $\lambda(t)$ is a given positive weighting function to keep the transition kernel $p_{0t}(\boldsymbol{x}_t|\boldsymbol{x}_0)$ to be a tractable Gaussian distribution. $\boldsymbol{\epsilon}_{\boldsymbol{X}} = \nabla_{\mathbf{X}}\mathrm{log~}p_{0t}(\mathbf{X}_t|\mathbf{X}_0)$ and $\boldsymbol{\epsilon}_{\boldsymbol{A}}=\nabla_{\mathbf{A}}\mathrm{log~}p_{0t}(\mathbf{A}_t|\mathbf{A}_0)$ are the sampled Gaussian noise. For the diffusion step $t$, we use a positional encoding method as described in~\citeauthor{ho2020denoising}\shortcite{ho2020denoising}, followed by two fully connected layers before feeding into the denoising network.

\subsection{Conditional Graph Denoising Network}

The architecture of our denoising network is shown in Figure~\ref{fig:framework} (a). 
The conditional denoising network processes a noisy graph represented by $\boldsymbol{G}_t = (X_t,A_t)$, alongside the regional fair-demand embedding $\boldsymbol{C}$ and the discrete graph $\boldsymbol{\bar A}$ as input conditions. The backbone network of our diffusion model integrates the graph transformer, as proposed by~\cite{dwivedi2020generalization}, with standard message-passing networks to form a Residual Hybrid Layer. This design enhances the model's ability to effectively process and integrate graph-based data. To incorporate time information, we normalize the time step to $[0,1]$ and treat it as a global feature to add noise. It outputs tensors $X$ and $A$ in different time intervals $t$, which respectively represent the optimization distribution over clean graphs.

\noindent\textbf{Graph Transformer Block.}
In each layer, the attention weights between every pair of nodes are calculated through self-attention. Notably, the calculation of attention weights for the propagation of information between nodes also incorporates the influence of edge features. 
\begin{align}
    & a_{i,j}=\mathrm{softmax}(\frac{(\tanh((W_{0}^{k,l}E_{i,j}^{l}))\cdot Q^{k,l})K^{k,l}}{\sqrt{d_{k}}}), 
    \\
    & M^{l+1}=\sum_{j=0}^{N-1}a_{ij}^{k,l}(\tanh(W_{1}^{k,l}(E_{i,j}^{l}))\cdot V^{k,l}) ,
\end{align} where $E_{i,j}$ is the corresponding image-like  tensor i.e. edge feature, \( Q \), \( K \), \( V \) are projected from the node features \( H_{i,j}^{(l)} \), \( H^{(l+1)} \in \mathbb{R}^{N \times d} \) represents the node outputs, \( W_0^{(l)} \) and \( W_1^{(l)} \) are learnable parameters, \(\tanh\) is the activation layer and $l$ is transformer layer. Our transformer layers also feature residual connections and layer normalization.

\noindent\textbf{Message Passing Block.} Since the graph transformer captures global graph information, in our task, there is also a need to focus on neighboring nodes within a 15-minute walking range. Therefore, we rely on the decoded discrete graph structure and employ a standard message-passing layer based on the Graph Convolutional Network (GCN) to aggregate local neighbor edge features. Then, we feed the node outputs $\Bar{H}^{(l+1)}$ into the \( \text{FFN}^{(l)} \), which comprises a multi-layer perceptron (MLP) and a normalization layer. Finally, we obtain the aggregated node output \( \bar{H}_{i,j}^{(l+1)} \in \mathbb{R}^{N \times d} \) and the corresponding edge features \( \bar{E}_{i,j}^{(l+1)} \in \mathbb{R}^{N \times N \times d} \).
\begin{align}
    & H_{i,j}^{l+1}=\mathrm{FFN}_0^l(\Bar{H}^{l+1})  ,
    \\
    & E_{i,j}^{l+1}=\mathrm{FFN}_1^l(\Bar{H}_i^{l+1}+\Bar{H}_j^{l+1}). 
\end{align}

Next, node and edge outputs from the graph transformer and message passing blocks are independently aggregated, then each fed into a fully connected layer. The resulting outputs, $\boldsymbol{\epsilon}_{\boldsymbol{\theta},\boldsymbol{X}}$ and $\boldsymbol{\epsilon}_{\boldsymbol{\theta},\boldsymbol{A}}$, are computed by aggregating data via skip connections from each residual hybrid layer, followed by processing through an MLP and three Conv1x1 blocks.


\noindent\textbf{Augmentation Module.}
We design a module to enhance node and edge features within graphs by incorporating statistical properties such as degree and centrality. Furthermore, using an m-step random walk derived from the discrete adjacency matrix, we develop $m$-step random walk matrices to employ the arrival probability vectors concatenated with original node features as enhanced node embeddings. Similarly, truncated shortest-path distances obtained from the same matrix are used to augment edge embeddings. Through node and edge augmentation, the network can comprehensively model the noisy topology structure from both node and edge perspectives at each diffusion step.

\begin{table*}[!ht]
\centering
\caption{Performance comparison of FAP-CD with baselines. The best results are in bold, and the second-best are underlined.}
\label{table:Overall Comparison}
\begin{tabular}{c|c|c|c|c|c|c}
\toprule
Model  & Life Service $\uparrow$       & Elderly Care $\uparrow$    & Diversity $\uparrow$   & Accessibility $\uparrow$   & Gini $\downarrow$  & Average $\uparrow$ \\
\midrule
Walking-based & $0.419$ & $0.223$ & $0.600$ & $0.167$ & $0.710$ & $0.140$ \\

ACA & $0.484_{\pm0.003}$ & $0.501_{\pm0.004}$ & $0.749_{\pm0.001}$ & $0.111_{\pm0.000}$ & $0.869_{\pm0.008}$ & $0.195_{\pm0.001}$ \\

GA & $0.445_{\pm0.004}$ & $0.393_{\pm0.005}$ & $0.645_{\pm0.015}$ & $0.102_{\pm0.013}$ & $0.479_{\pm0.005}$ & $0.221_{\pm0.003}$ \\

DRF & $0.449_{\pm0.098}$ & $0.604_{\pm0.083}$ & $0.728_{\pm0.053}$ & $0.123_{\pm0.024}$ & $0.376_{\pm0.139}$ & $0.306_{\pm0.024}$ \\

VGAE & $0.180_{\pm0.020}$ & $0.263_{\pm0.125}$ & $0.218_{\pm0.001}$ & $0.356_{\pm0.086}$ & $0.701_{\pm0.033}$ & $0.063_{\pm0.040}$ \\

GraphRNN & $0.442_{\pm0.022}$ & $0.546_{\pm0.035}$ & $0.503_{\pm0.061}$ & $0.361_{\pm0.004}$ & $0.387_{\pm0.016}$ & $0.293_{\pm0.021}$ \\

CondGEN & $\underline{0.643_{\pm0.092}}$ & $0.667_{\pm0.059}$ & $0.563_{\pm0.010}$ & $0.345_{\pm0.041}$ & $0.323_{\pm0.029}$ & $0.379_{\pm0.035}$ \\

DDPM & $0.564_{\pm0.028}$ & $\underline{0.689_{\pm0.030}}$ & $\underline{0.820_{\pm0.015}}$ & $\underline{0.402_{\pm0.020}}$ & $0.394_{\pm0.027}$ & $\underline{0.416_{\pm0.014}}$ \\

EDGE & $0.431_{\pm0.002}$ & $0.590_{\pm0.068}$ & $0.667_{\pm0.055}$ & $0.234_{\pm0.115}$ & $\underline{0.327_{\pm0.003}}$ & $0.341_{\pm0.051}$ \\

\midrule
FAP-CD (Ours) & $\bf{0.888_{\pm0.015}}$ & $\bf{0.923_{\pm0.031}}$ & $\bf{0.843_{\pm0.004}}$ & $\bf{0.520_{\pm0.103}}$ & $\bf{0.232_{\pm0.013}}$ & $\bf{0.588_{\pm0.028}}$ \\

\bottomrule
\end{tabular}
\end{table*}

\section{Experiments}

\subsection{Experimental Setup}
\textbf{Data Description.}
We develop our model using real-world datasets, focusing on Beijing. We obtain geographic information from OpenStreetMap, including road networks and administrative divisions. Additionally, we source 149,049 Points of Interest (POIs) within the city from Baidu Maps\footnote{https://map.baidu.com}, categorized into 11 types such as food, medical services, etc. We also gather data on 568 senior meal programs and 1,472 senior care centers from\footnote{https://www.beijingweilao.cn}. In collaboration with the real estate agency\footnote{https://m.lianjia.com}, we acquire comprehensive data on 8,201 residential areas, detailing housing prices, property fees, community populations, and elderly populations over age 60.

\noindent\textbf{Evaluation Metrics.}
To comprehensively evaluate our model, we employ a set of evaluation metrics that take into account facility types, spatial distribution and fairness, which provide a multi-dimensional view of the effectiveness and equity of the allocations. The key evaluation metrics include: \textbf{Efficiency}$=$$\frac{1}{|R| \cdot |K|} \sum\limits_{i \in R} \sum\limits_{k \in K} \sum\limits_{h \in H_i} \frac{\max_{n \in N_{ik}} A_{in,h}}{|N_{ik}|}$, 
\textbf{Diversity}$=$$\frac{1}{|R|} \sum\limits_{i \in R} \left( \frac{\left| \{ C_{i,j} \mid j \in S_i \text{ and } C_{i,j} \neq R_i\} \right|}{N - 1} \right)$, \textbf{Accessibility}$=$$\frac{1}{|R|} \sum\limits_{i \in R}\sum\limits_{h \in H_i} \frac{\max\limits_{n \in N_{ik}} A_{in,h}}{P_i\cdot u}$, 
\textbf{Gini}$=$$1 - \frac{2\cdot\sum\limits_{i=1}^{N} \sum\limits_{j=1}^{i} X_{(j)}}{N\cdot\sum\limits_{i=1}^{N} X_{i}}$. In this study, \textbf{Efficiency} refers to the distribution efficiency of life service and elderly care facilities, as measured by the Life Service and Elderly Care metrics. The \textbf{Average} metric assesses the overall performance on all metrics, where the Gini metric is negated.

\noindent\textbf{Baseline Models.} We compare our proposed FAP-CD with two classes of methods. The first type is 
\textbf{1) Traditional methods}: Walking-based~\cite{song2024supply}. The Ant Colony Algorithm (ACA)~\cite{yang2007parallel} generates an initial set of feasible metro lines and then optimizes community planning through crossover and mutation. The Genetic Algorithm (GA)~\cite{jensen2019graph} guides station selection with a pheromone-based rule and updates edge pheromones to optimize community planning. DRF~\cite{ghodsi2011dominant} dynamically allocates resources to the user (grid region) with the lowest dominant share until the user's demand cannot be met. And the other \textbf{2) Deep generative models}: VGAE~\cite{kipf2016variational} uses GCN to encode nodes into a latent space and then reconstructs the graph by decoding the adjacency matrix from the node embeddings. GraphRNN~\cite{you2018graphrnn} sequentially generates graphs by using recurrent neural networks to model the dependencies between nodes and edges. CondGEN~\cite{yang2019conditional} is a variational graph generative model that integrates flexible structures while maintaining permutation invariance. DDPM~\cite{ho2020denoising, wen2023diffstg} uses the UGnet framework to build a diffusion process applied to node representations and image-like adjacency matrices. EDGE~\cite{chen2023efficient} leverages graph sparsity during the diffusion process by focusing on a small subset of nodes, improving efficiency and addressing the challenges of large-scale graph generation. In this baseline, we use node category sequences as guidance instead of node degrees.

\noindent\textbf{Implementation Details.} 
We conduct experiments using two Nvidia A40 GPUs, configuring 200 diffusion steps and employing a 3rd-order DPM-Solver~\cite{lu2022dpm} along with a cosine noise scheduler~\cite{nichol2021improved}.

\subsection{Experimental Results}

\subsubsection{Overall Comparison.}
Table~\ref{table:Overall Comparison} shows the overall comparison results in terms of all evaluation metrics in Beijing. Our observations based on the results are as follows. First, the AFCs planning generated by FAP-CD achieves the best realism in terms of both facility distribution efficiency and equity. Specifically, compared with the best baseline, the \textit{Accessibility} is improved by over 29\% and \textit{Gini} index reduced over 28\%, respectively. Moreover, our model outperforms all baselines on average by a large margin. Second, We discover that when categorizing facility allocations, traditional resource allocation models, particularly DRF, exhibit superior diversity. This superiority arises because the facility allocations in methods AC and GA are guided by a demand-driven pheromone update mechanism and fitness functions, respectively, and are also influenced by stochastic factors.

Our analysis further reveals that the overall performance of the EDGE is inferior to that of DDPM and our model. This indicates that while graph generation guided by node categorization presents significant advantages in terms of graph discretization and time efficiency, relying solely on a denoising network composed of MLPs and Gated Recurrent Units (GRUs) to extract neighborhood and global spatial information from the graph does not optimize graph generation. This underscores the relative advantages of our hybrid message passing and augmentation strategy.

\begin{figure}[!t]
	\centering
    \includegraphics[width=0.42\textwidth,height=0.25\textwidth]{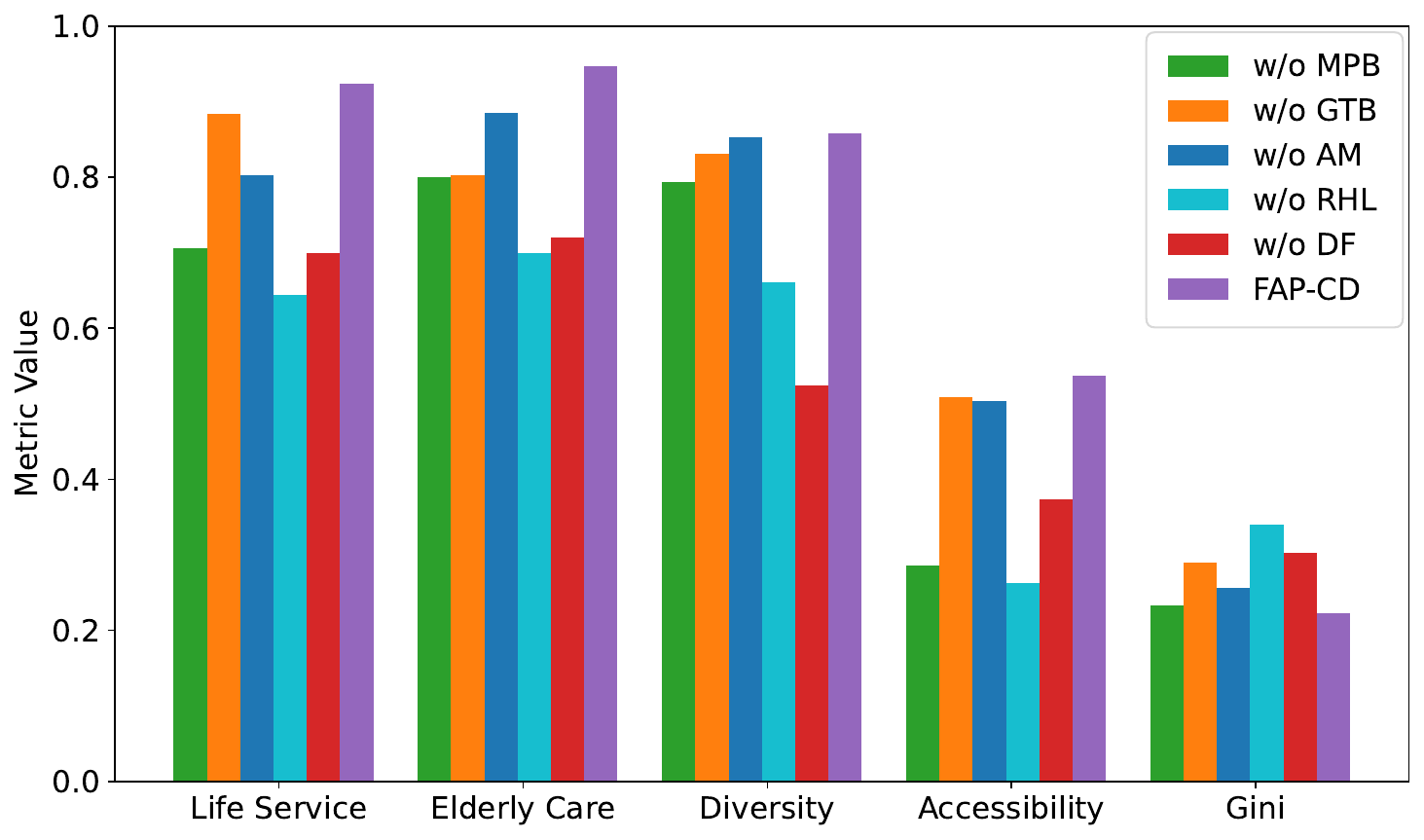}
    \caption{Performance Comparison in the Ablation Study.}
	\label{fig:ablation}
	\vspace{-2ex}
\end{figure}

\begin{figure}[ht]
	\centering
	\subfigure[Hidden size]{
		\includegraphics[width=0.48\linewidth]{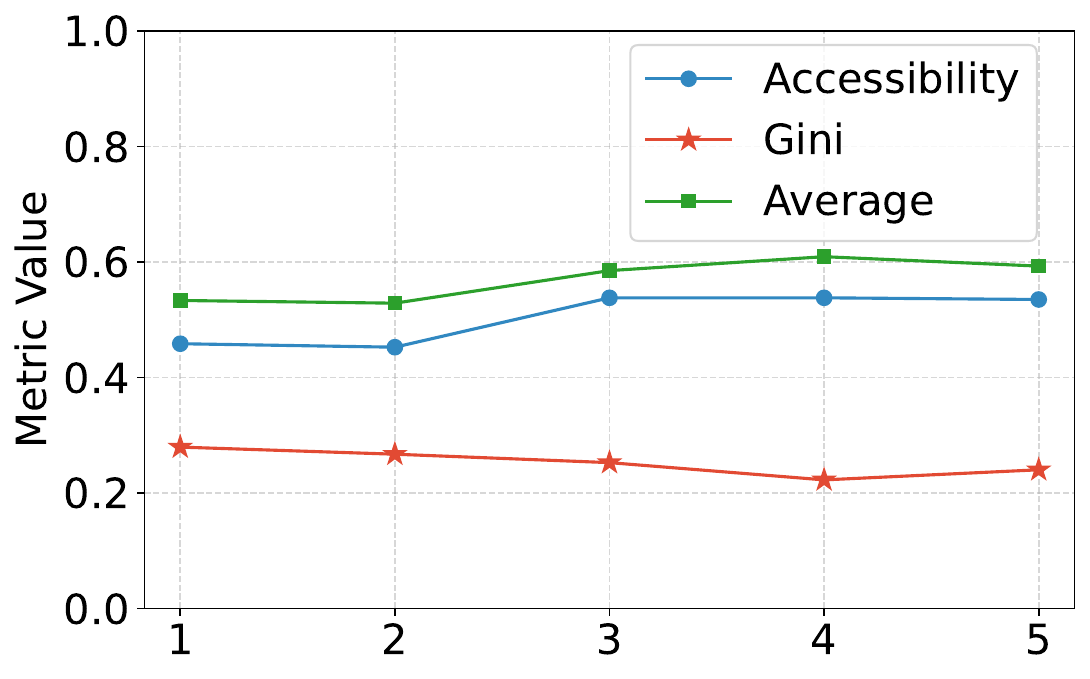}}
	\subfigure[Layer number]{
		\includegraphics[width=0.48\linewidth]{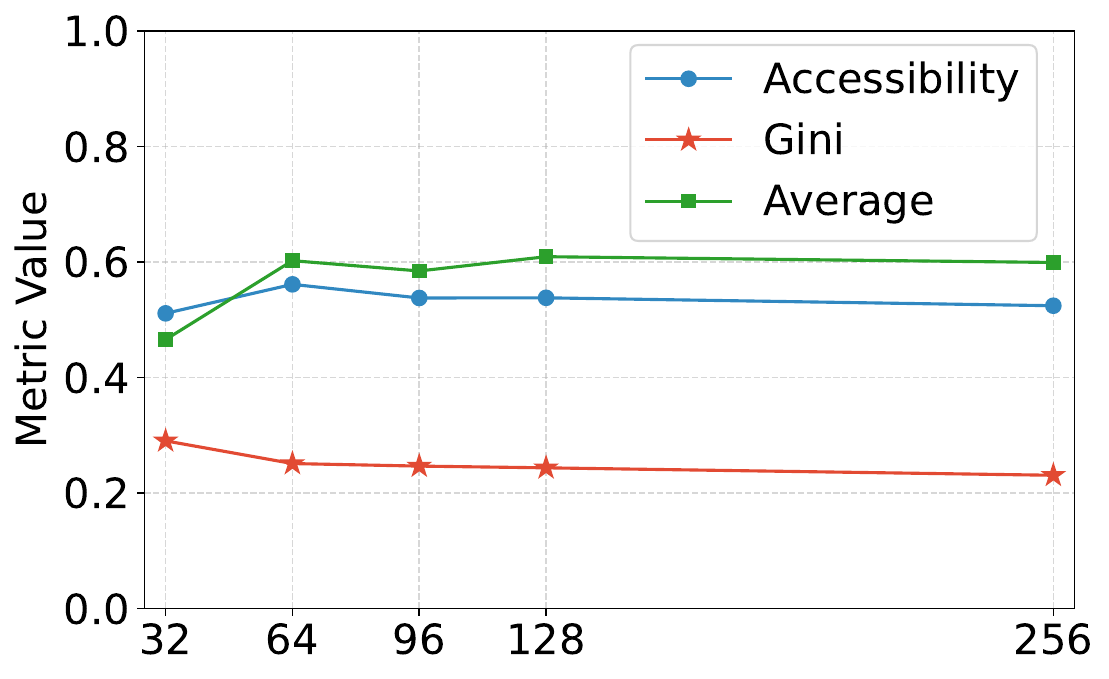}}
    \caption{Analysis of Hyperparameters.}
	\label{fig:hyper}
\end{figure}
\subsubsection{Ablation Study.}
We conduct an ablation study to validate the efficacy of the Residual Hybrid Layer. Results in Figure~\ref{fig:ablation} indicate that incorporating the residual hybrid module (w/o RHL) within the denoising model enhances AFCs generation performance by up to 105.1\% according to metric \textit{accessibility}. Additionally, ablation experiments are conducted separately on the graph transformer block (w/o GTB) and message passing block (w/o MPB) within the denoising network, revealing that removing either mechanism causes performance degradation to varying degrees. We also evaluate the impact of the node and edge attribute enhancement module (w/o AM) within the SDE-based diffusion model. As shown in Figure~\ref{fig:ablation}, this design yields up to a 13.1\% improvement in performance according to metric \textit{Gini}, marking the most significant enhancement observed.

\subsubsection{Hyperparameters.}
We next conduct experiments on various hyperparameters, including the number of residual hybrid layers in the conditional graph denoising network and the dimension of hidden variables. As illustrated in Figure~\ref{fig:hyper} (a), the results show that a shallow and appropriate number of layers (three in this case) is crucial for FAP-CD, as overly deep networks tend to overfit, leading to a decline in performance. Figure~\ref{fig:hyper} (b) examines the impact of hidden variable dimension size. The results indicate that an optimal hidden size (e.g., 128) enhances model performance, whereas excessively large dimensions (e.g., 256) do not improve performance and increase the computational burden.

\subsubsection{Visualization of Generated AFC Configurations.}
\begin{figure}[!tb]
\centering
\includegraphics[width=0.9\linewidth]{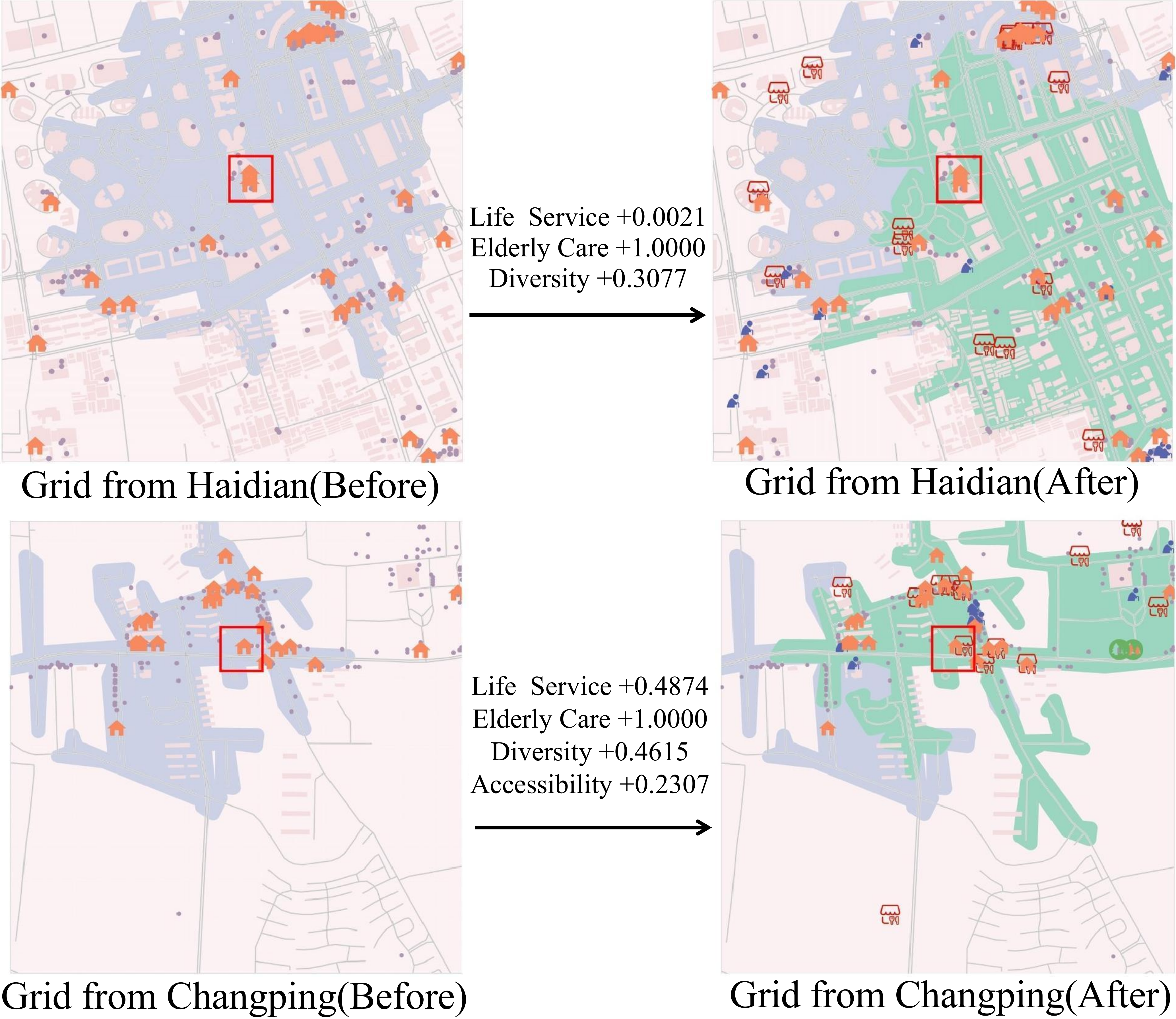}
\caption{Comparison of Generated and Original AFCs.}
\label{fig:case study}
\end{figure}

Figure~\ref{fig:case study} presents a visual comparison between the original and generated AFCs in Beijing, analyzing the facility distribution across various land parcels. The results show that the generated plan not only enhances community diversity but also significantly improves the efficiency of aging-friendly facility distribution. Furthermore, our model optimizes the spatial arrangement of these facilities. 
For example, in the lower grid, areas highlighted in green as 15-minute walkable zones and outlined by red boxes now enjoy improved access to previously hard-to-reach facilities, such as hospitals and parks. This enhancement is facilitated by reconfigured roadways, including major urban renewal projects like subways and overpasses. Thus, FAP-CD offers urban planning experts valuable insights for facility planning and clearly demonstrates its potential to optimize spatial allocations.

\section{Related Work}

\noindent\textbf{Diffusion Models.} Diffusion models, as an emerging generative technique, have demonstrated exceptional capabilities in fields such as computer vision~\cite{dhariwal2021diffusion}, natural language processing~\cite{li2022diffusion}, and graph generation~\cite{niu2020permutation, vignac2022digress}.
These models operate through two main phases: an initial noise addition phase and a reverse denoising phase that restores noisy data to complex distributions. The most pertinent works involve the application of diffusion models to graph generation tasks~\cite{niu2020permutation, vignac2022digress, huang2023conditional}. For instance, \citeauthor{niu2020permutation}\shortcite{niu2020permutation} used score-based models with Gaussian noise for permutation-invariant graph, while \citeauthor{jo2022score}\shortcite{jo2022score} and recent studies ~\cite{vignac2022digress, haefeli2022diffusion} explored the potential of diffusion models for generating node and edge-enhanced graphs and using multinomial noise for discrete models that preserve graph sparsity and enhance quality. Moreover, incorporating additional information as conditioning cues is crucial for targeted generation. For example, 
~\citeauthor{zhou2023towards}\shortcite{zhou2023towards} designed a novel diffusion process to generate city mobility flows, which are guided by a volume estimator and influenced by regional features. 

Inspired by the significant success of diffusion models, our study integrates age-friendly needs and urban region characteristics into our model to optimize AFCs planning. This study explores the capabilities of graph diffusion models in strategic planning for AFCs, aiming not only to develop configurations for community revitalization but also to provide insights into optimal geographic locations.

\noindent\textbf{Automated Planning Generation.} Traditional urban planning methods have encountered many challenges in practice, including high cost, low efficiency, lack of skilled facilitators, and low interest to participate~\cite{abas2023systematic}. Recently, the remarkable success of deep generative learning has led researcher about automated planning generation to improve the efficiency of urban planning. For example,~\citeauthor{wang2023human}\shortcite{wang2023human} eveloped an instruction-based hierarchical planner (IHPlanner), which automatically captured the spatial hierarchy and planning dependencies between urban functional zones and land use allocation based on human instructions.~\citeauthor{zheng2023spatial}\shortcite{zheng2023spatial} proposed a RL-based urban planning model that trained a value network to enhance spatial planning quality focusing on the completeness of the 15-minute city concept~\cite{moreno2021introducing}, with the aim of improving the efficiency of transportation, services, and ecological systems in urban regions. However, these plans are either completed based on single human instructions or remain inherently time-consuming. Furthermore, they lack scalability in age-friendly service facilities and do not address issues of planning efficiency and fairness. Compared to these efforts, the FAP-CD model is more advanced in automation and the facilitation of age-friendly urban planning.

\section{Conclusion}

In this work, we propose a novel conditional graph diffusion model FAP-CD, which transforms the planning issue into a problem of graph generation by constructing a detailed walking graph of spatial relationships among facilities within a region. This model leverages fair-demand guidance enhancement and discrete graph structure adjustments along with a finely tuned graph noise prediction network to generate AFCs planning during the reverse denoising process. Extensive experiments and case studies validate FAP-CD's effectiveness, enhancing efficient and equitable age-friendly urban renewal planning and supporting sustainable urban development amid rapid global aging.

\section{Appendix}

\textbf{A. Details of Datasets}

\noindent\textbf{Urban Geospatial Dataset.} 
The urban geospatial dataset\footnote{openstreetmap.org} employed in this study encompasses a comprehensive collection of geographic information within the Beijing metropolitan area, including pedestrian road networks, building footprints, and township-level administrative divisions. The pedestrian road networks are used to generate service areas of 15-minute walking life circles surrounding residential areas. The administrative division data are organized at the township and community levels of Beijing's governance boundaries. The building footprint data outlines the spatial extent and structure of individual buildings.

\noindent\textbf{Senior Service Dataset.} The senior service dataset encompasses POI data specifically focused on elderly service stations\footnote{https://map.baidu.com} and senior meals canteens\footnote{https://www.beijingweilao.cn} within Beijing. For the elderly service stations, data points include the station's level, the range of services provided, insurance accreditation status, and details about the establishment of medical rooms within the facilities. Regarding the senior meal canteens, the dataset provides essential information such as the addresses, operating hours, and pricing of the food offered.

\begin{figure}[hbtp]
	\centering
	\subfigure[The distribution of facility categories.]{
		\includegraphics[width=0.45\linewidth]{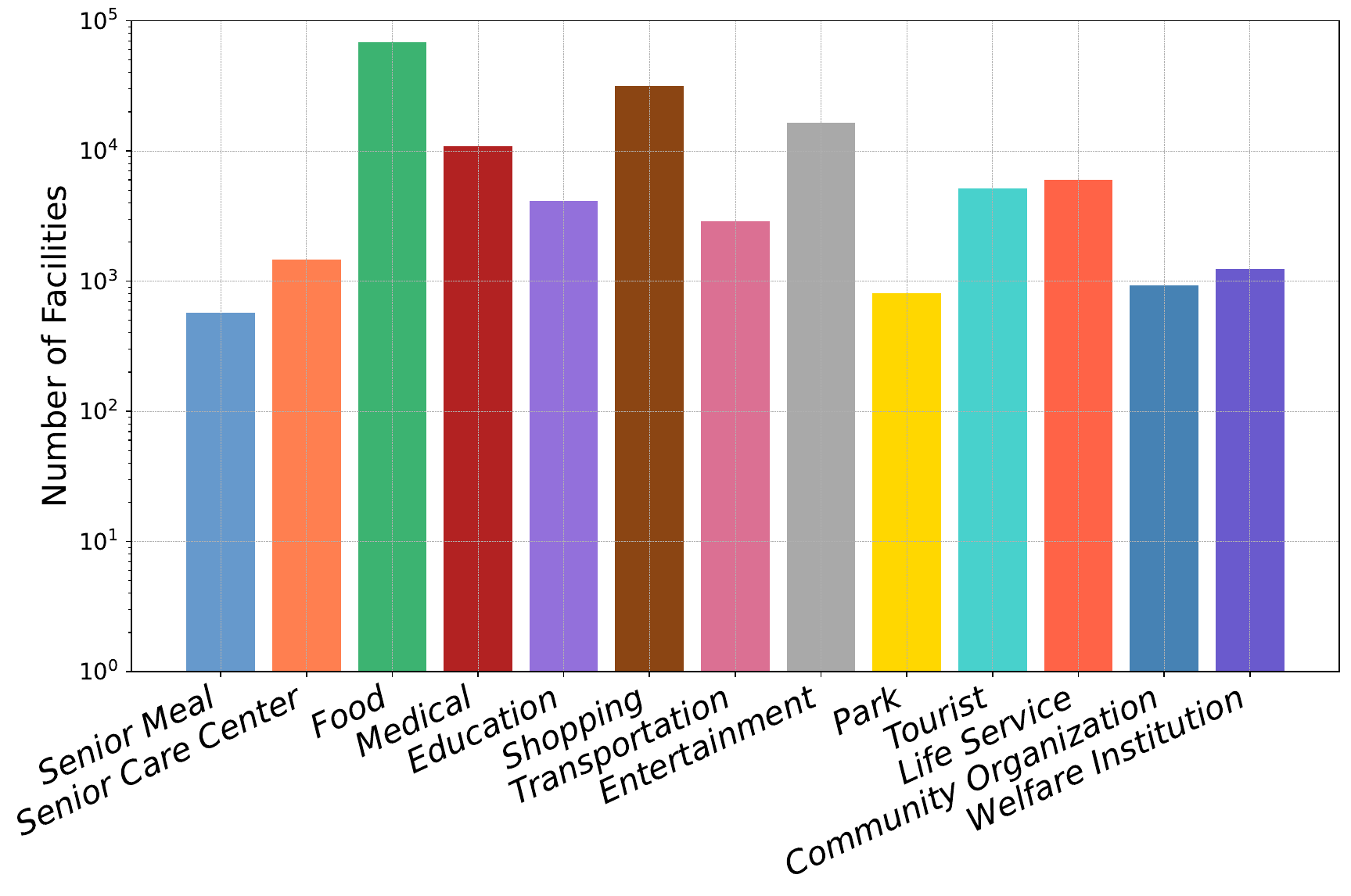}}
	\subfigure[Average street-level house prices in Beijing.]{
		\includegraphics[width=0.45\linewidth]{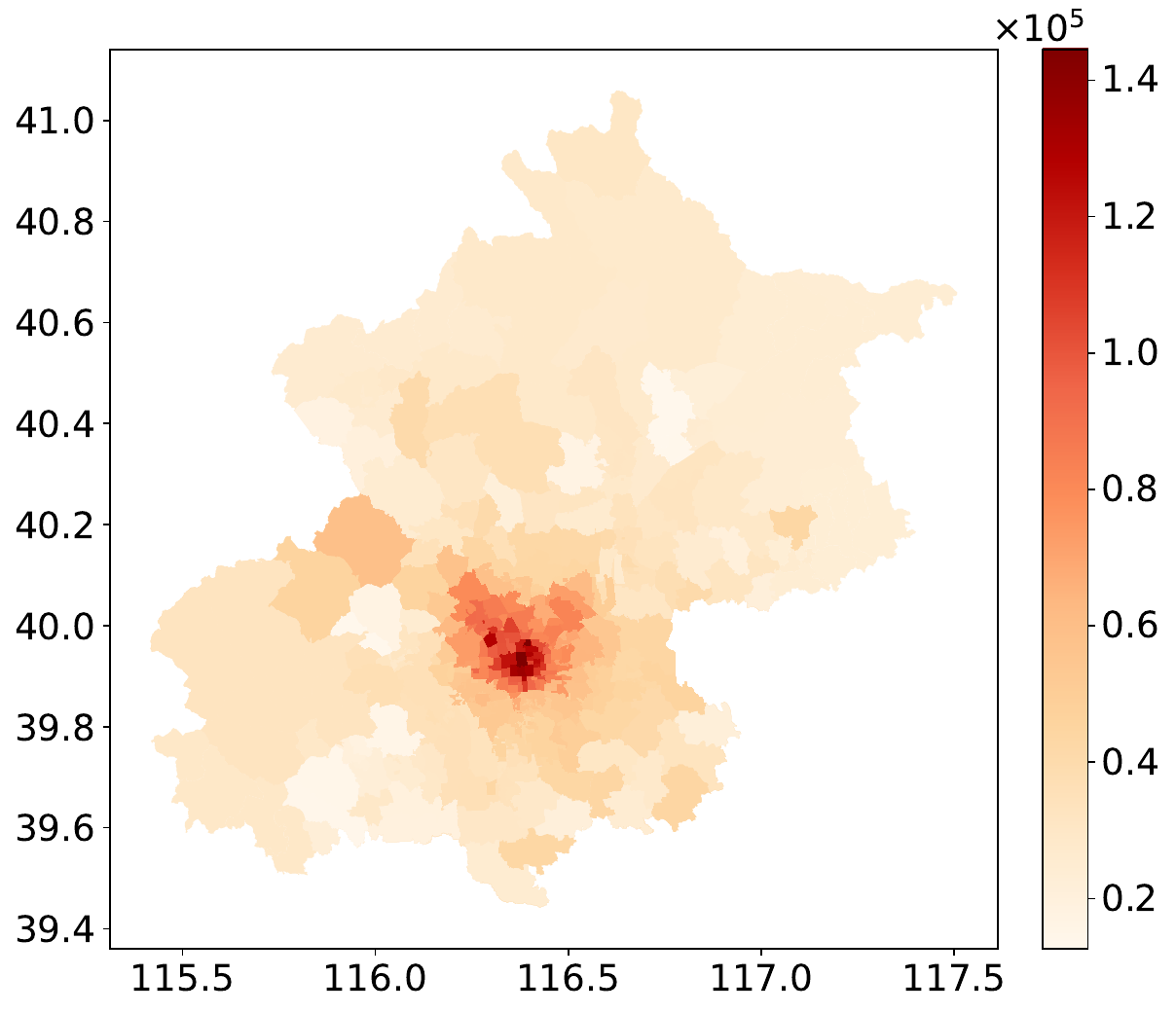}}
        \subfigure[The elderly population of the streets in Beijing.]{
		\includegraphics[width=0.45\linewidth]{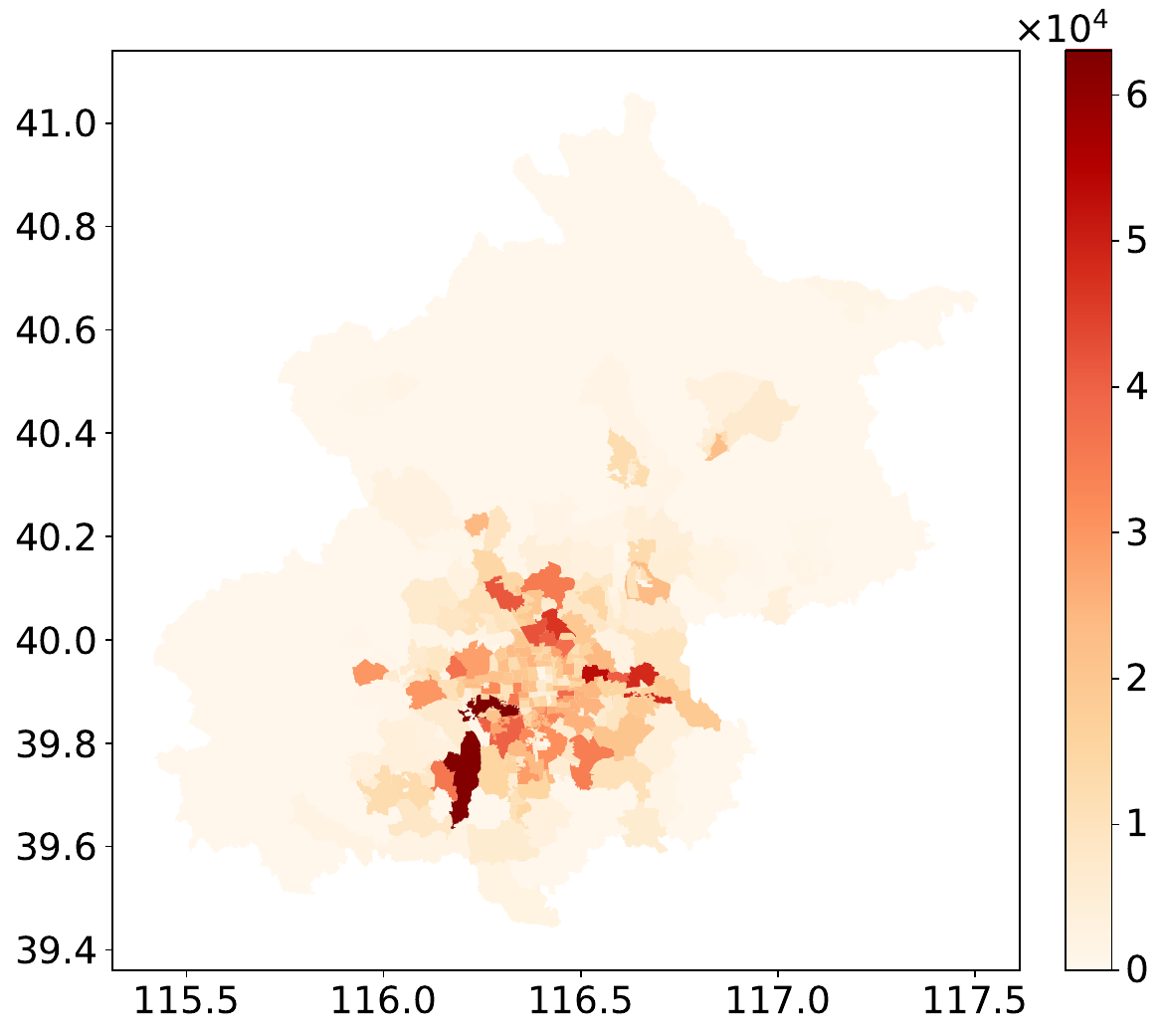}}
        \subfigure[Number of elderly care facilities in Beijing.]{
		\includegraphics[width=0.45\linewidth]{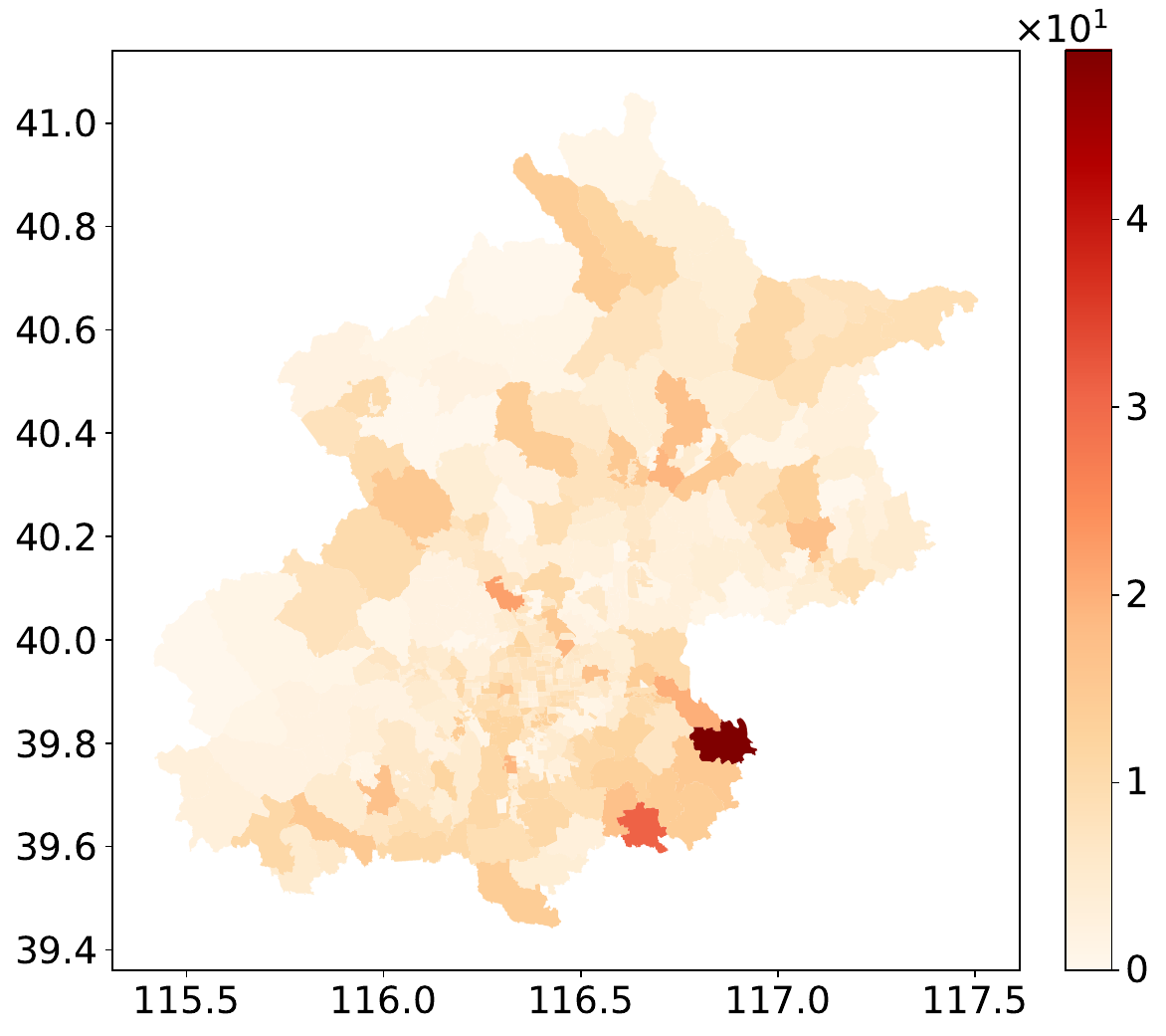}}
	\caption{Details of datasets in Beijing.}
	\label{fig:facilities}25
\end{figure}

\noindent\textbf{Residential Area Dataset.} 
The residential area dataset comprises the residential communities within Beijing. The basic information includes fundamental details about each residential area, such as the name, address, geographical coordinates, house prices, total numbers of buildings, and housing units. Additionally, we integrate data from the 2020 Seventh National Population Census of China\footnote{https://nj.tjj.beijing.gov.cn/tjnj/rkpc-2020/indexch.htm} to obtain more precise figures for the resident population and the elderly population aged 60 and above within each residential area, which allows for a more accurate analysis of demographic trends and the specific needs of different communities. In Figure~\ref{fig:facilities}, we visualize the distribution of facility categories and assess the per capita availability throughout Beijing.

Our experimental dataset contains 466 grid regions with residential areas. The number of facilities in these grids ranges from a minimum of 10 to a maximum of 399, including residences.

\noindent\textbf{B. Implementation Details}\\
In our experiments, we set the number of diffusion steps to 200, adopting a 3st-order DPM-Solver~\cite{lu2022dpm} and utilizing a cosine noise scheduler as proposed by~\citeauthor{nichol2021improved}\shortcite{nichol2021improved}. Our model is trained for 1000 epochs, using four residual layers in the graph denoising network with a hidden embedding dimension of 128. The Adam optimizer is employed with an initial learning rate of 3e-4, following a decay schedule of 1e-2. The batch size for the graph denoising network is set to 8. The number of random walks in the augmentation module is set to 20. For the 
\textit{fair-demand module}, the training is conducted for 60 epochs with a batch size of 6 and a learning rate of 1e-5.

To ensure a fair comparison, we use node embeddings from our pretrained model as the initial node features for VGAE and GraphRNN, and a MLP decoder with inner product computation for edge existence to  reconstruct the graph structure.

We provide a detailed description of the experimental setup in the baseline EDGE, which refers to the experimental setup for hierarchical generation with EDGE in Appendix E to generate graphs with node attributes. Specifically, node attributes are represented as one-hot vectors, we have a matrix $\mathbf{X}\in\{0,1\}^{N \times C_{node}}$, while adjacency matrix is described by the matrix  $\mathbf{A}\in\{0,1\}^{N \times N}$ to indicate whether the edge exists or not. In this context, $C_{node}$ denotes the number of classes for node types. We consider the following joint model:
\begin{equation}
    p(\mathbf{X}, \mathbf{A}) = p(\mathbf{X}) p(\mathbf{A} | \mathbf{X})
\end{equation}

We first sample node attributes, employing the node sequence length $C_{node}$ instead of the node degree sequence modeling. For $p(A|X)$, we apply the EDGE framework, incorporating node features from X during both the training and generation phases. Subsequently, the graph structure is sampled using EDGE, conditioned on node attributes. 

\noindent\textbf{C. Hyperparameters for SDE}\\
We apply the variance-preserving SDE (VPSDE) introduced by~\cite{song2020score} for graph data $\boldsymbol{G}_t$, whose discrete-time form is a denoising diffusion probabilistic model. Considering the linear scaling function $\beta(t) = \bar{\beta}_{min}+t(\bar{\beta}_{max}-\bar{\beta}_{min})$ for $t\in[0,T]$, and $f(t) = -\frac12\beta(t), g(t) = \sqrt{\beta(t)}$, we derive the SDE from equation 2 presented in the main text as:
\begin{equation}
\label{Eq:eq1}
\mathrm{d}\boldsymbol{G}_t=-\frac12\beta(t)\boldsymbol{G}_t\mathrm{d}t+\sqrt{\beta(t)}\mathrm{d}\boldsymbol{w}_t ,
\end{equation}
where $\beta(t)$ follows a monotonically decreasing cosine schedule across different timesteps $t$ and controls the noise scale. The values for $\bar{\beta}_{max}$ and $\bar{\beta}_{min}$ are set to 20 and 0.1, respectively. The parameters of VPSDE are used without further tuning, maintaining a small signal-to-noise ratio at $\boldsymbol{G}_t$.

\noindent\textbf{D. Details of Evaluation Metrics}\\
The metrics primarily assess the efficiency, spatial layout, and fairness of the generated plans. 
Specifically, to evaluate the coverage of daily life needs and the elderly-friendliness of the plans, we calculate the efficiency of both elderly care and life services. 
\textbf{Efficiency} measures the extent to which facilities cover residential nodes and assesses the overall effectiveness of the plan by averaging the coverage rates across sample grids. 
To assess the diversity of facility types within the plan, we introduce the \textbf{Diversity} metric, which evaluates the variety of facilities across different grids. 
\textbf{Accessibility}~\cite{talen1998assessing} measures the convenience and rationality of the spatial distribution of facilities, taking into account the population size and the accessibility of facilities. 
Moreover, to comprehensively assess the fairness of the plan, we use the \textbf{Gini} coefficient~\cite{jang2017assessing}, which combines the efficiency of Accessibility, elderly care, and life services to evaluate the equity of facility layout across different grids. 
After aligning the directionality of these five metrics, we calculated a comprehensive \textbf{Average} metric, providing an overall assessment of the facility allocation plan’s performance across multiple dimensions.
\begin{small}
\begin{align}
    \centering
    & \text{Efficiency} = \frac{1}{|R| \cdot |K|} \sum\limits_{i \in R} \sum\limits_{k \in K} \sum\limits_{h \in H_i} \frac{\max_{n \in N_{ik}} A_{in,h}}{|N_{ik}|} ,
    \\
    & \text{Diversity} = \frac{1}{|R|} \sum\limits_{i \in R} \left( \frac{\left| \{ C_{i,j} \mid j \in K_i \text{ and } C_{i,j} \neq Re_i\} \right|}{N - 1} \right) ,
    \\
    & \text{Accessibility} = \frac{1}{|R|} \sum\limits_{i \in R}\sum\limits_{h \in H_i} \frac{\max\limits_{n \in N_{ik}} A_{in,h}}{P_i\cdot u} ,
    \\
    & \text{Gini} = 1 - \frac{2\cdot\sum\limits_{i=1}^{N} \sum\limits_{j=1}^{i} X_{(j)}}{N\cdot\sum\limits_{i=1}^{N} X_{i}} ,
    \label{eq:metric}
\end{align}
\end{small}where $R$ is the set of grids, $N$ is the total number of grids, and $K$ is the set of facility types, which in our case includes 14 types. $H_i$ represents the residential indices in grid region $r_i$, while $N_{ik}$ denotes the indices of facility type $k$ within the same grid. $A$ is the adjacency matrix, $C_{ij}$ indicates the category of node $j$, and $Re_i$ represents the index of the residential category. $P_i$ represents the total population in grid $r_i$, with $u$ as the population unit (in this study, per thousand people). $X_i$ is the composite metric value for the grid, integrating \textit{life service}, \textit{elderly care}, and \textit{accessibility} metrics, and $X_{(i)}$ is the sorted value of $X_i$ in ascending order.

\begin{figure}[!tb]
\centering
\includegraphics[width=0.95\linewidth]{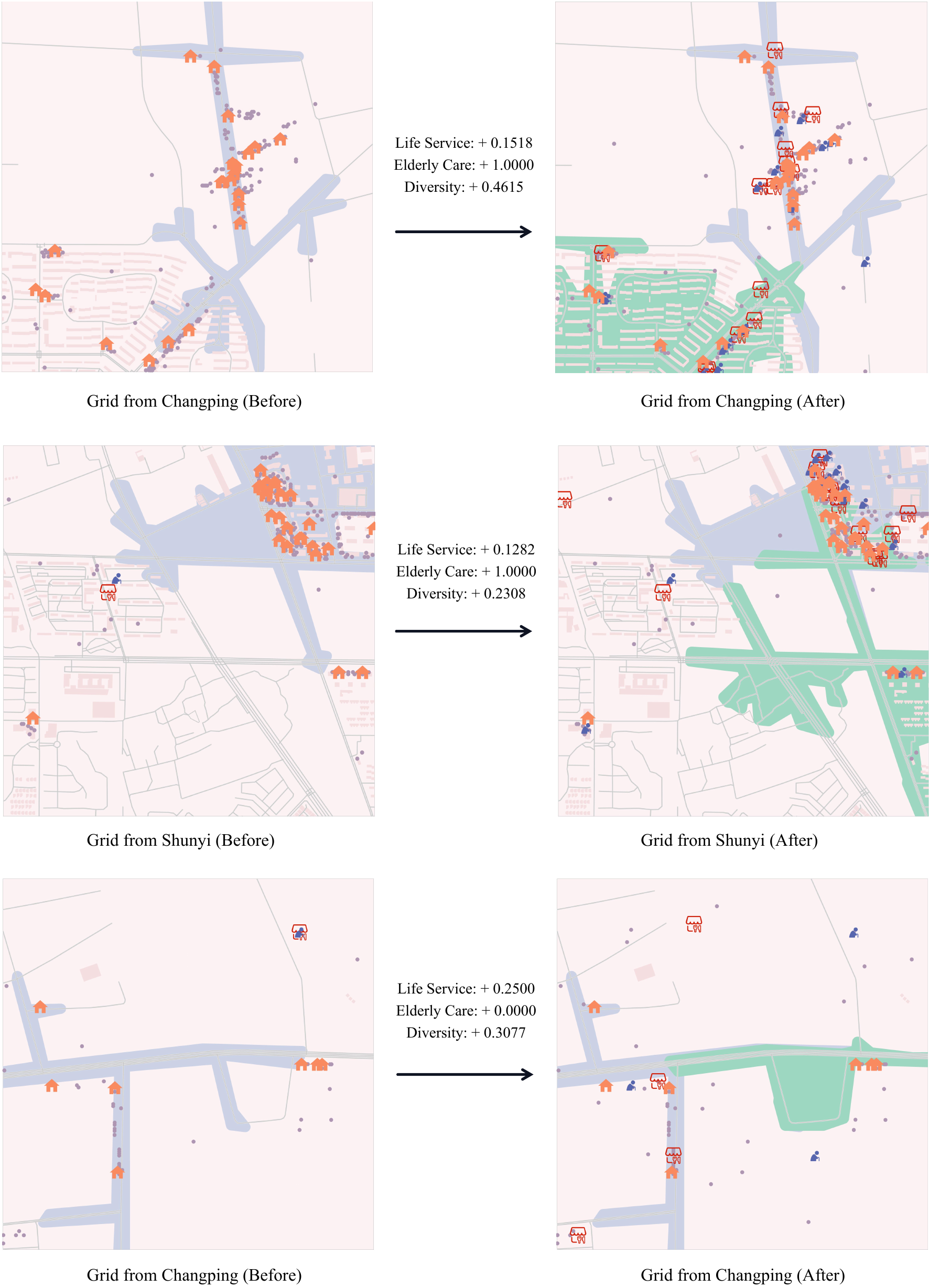}
\caption{Visual comparison pairs of generated AFCs.} \label{fig:Generated AFCs}
\end{figure}

\noindent\textbf{E. Visualization of the Generated AFCs Comparison}\\
Figure~\ref{fig:Generated AFCs} visualizes the comparison pairs of AFCs planning schemes across different grid areas in Beijing. The results indicate that the generated AFCs planning not only enhances community diversity but also significantly improves the efficiency of facility distribution.
In the first and third grid regions, residential zones exhibit a dispersed pattern along the roadways but lack elderly care facilities. The distribution generated by our model significantly enhances the allocation efficiency of aging-friendly facilities, while also improving the efficiency of life services and increasing the overall diversity within these areas. In the second grid, which contains over 20 densely clustered residential zones, the model similarly demonstrates improvements in both the efficiency and diversity of facility allocation. Furthermore, it is noteworthy that in all three case regions, the generated aging-friendly facilities are predominantly located near residential zones.

\nobibliography*
\bibliography{aaai25}

\end{document}